# ANOMALOUS ENTITY DETECTION USING A CASCADE OF DEEP LEARNING MODELS

*Hamza Riaz, Muhammad Uzair and Habib Ullah*
COMSATS University Islamabad, Wah Campus, Pakistan
ihamzariaz@gmail.com, {uzair,habib}@ciitwah.edu.pk

## ABSTRACT

Human actions that do not conform to usual behavior are considered as anomalous and such actors are called anomalous entities. Detection of anomalous entities using visual data is a challenging problem in computer vision. This paper presents a new approach to detect anomalous entities in complex situations of examination halls. The proposed method uses a cascade of deep convolutional neural network models. In the first stage, we apply a pretrained model of human pose estimation on frames of videos to extract key feature points of body. Patches extracted from each key point are utilized in the second stage to build a densely connected deep convolutional neural network model for detecting anomalous entities. For experiments we collect a video database of students undertaking examination in a hall. Our results show that the proposed method can detect anomalous entities and warrant unusual behavior with high accuracy.

## 1. INTRODUCTION

Surveillance cameras play a key role in providing safety in various indoor and outdoor human environments such as roads, banks, hospitals, airports, and shopping malls etc. [1][2]. Due to the exponential increase of visual data, it is almost impossible to analyze surveillance videos manually to monitor for abnormal situations. Therefore, machine and deep learning techniques are on the rise to build automatic systems for human behavior analysis in video data. However, recent literature indicates that automatic human behavior analysis especially anomalous entity detection is still an unsolved problem [3-6].

General outdoor and indoor applications of anomalous entity detection have major requirements such as accuracy, speed, and generalization capability. Recently, many machine learning methods of anomaly detection have been proposed for various applications like pedestrian flows, crowd flows, security surveillance at airports and train stations [1-6]. However, designing a single global machine learning model is challenging because the anomaly detection problem commonly involves multiple aspects of complex human behavior even when studied under the same application context.

Here we consider the problem of detecting abnormal behavior such as cheating in examination rooms. Figure 1 illustrates the basic building blocks of our proposed system. Our method uses a cascade of two robust deep convolutional neural networks models including a real time 2D human pose estimator known as the OpenPose model [7], and a densely connected convolutional neural network model known as the DenseNet [8]. The motivation behind our idea of exploring OpenPose comes from the observation that the information of poses and joints presents strong cues for understanding human actions. OpenPose also works robustly in indoor environments such as a classroom or examination hall. The extracted OpenPose patches of each joint are fed into a DenseNet which is a powerful deep neural network with key advantages in terms of accuracy, speed, parameters efficiency, and understanding of the model [8].

For experiments and tests we record numerous videos collected from volunteer students depicting examination setup in a hall. Overall, we labeled 157245 patches of $32\times32\times3$ resolution with two classes corresponding to all the key points detected by the OpenPose. We observed that using only the patch corresponding to the head key point gave satisfactory results. Therefore, we only consider 30951 patches of faces for training of DenseNet. Our approach involves the training of DenseNet from the scratch which helps to extract deeper features of input patches like direction of eyes and movements of heads for each candidate in the exam room. In the end, we perform inference on our trained model on 10,318 test images and evaluate results in terms of precision and sensitivity.

## 3. PROPOSED APPROACH

The literature review indicates that anomaly detection is a hard problem to solve due to the occurrence of different complex scenarios. For our system, we focus on indoor examination rooms. The main goal is to automatically label anomalous frames and present a summarized report to a human reviewer. Figure 1 illustrates the basic building blocks of our proposed system.

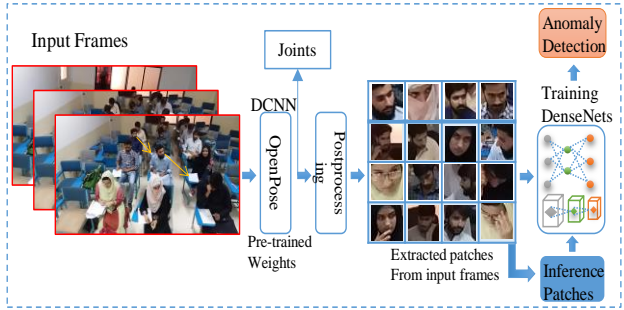

**Figure 1** Illustration of the proposed method.

### 3.1. Development of Dataset
One of the main tasks in our idea was collecting dataset and converting it to a meaningful form. After getting videos, we converted them into frames. For the initial training we used one video and get 2546 frames. Another task was to label each normal and abnormal action of exam participants. Therefore, we designed a MATLAB based tool for labeling each individual person in every frame. The results of pretrained OpenPose model helped the annotation tool to extract and label patches of faces, both elbows and hands.

### 3.2. 2D Human Pose Estimation
This paper uses a real time 2D pose estimation model called OpenPose [7] which gives potential to a computer to understand human actions in videos. The OpenPose model can detect 18 different body joints of humans. These 18 points are essential for behavior understanding but to build our intelligent invigilator, five joints including face, elbows, and hands can be considered. However, this paper is exploring face patches as a proof of concept.

### 3.3. Postprocessing
In the postprocessing, the proposed method extracts temporal patches of resolution 32×32×3. At this point, our special GUI tool assist users to label such complicated dataset robustly and accurately. We then scale and normalize the patches by using means and standard deviations of each patch.

### 3.4. DenseNet Model
The idea of DenseNet [8] originated from the ResNet [9]. The DenseNet's unique connectivity of layers, which is also known as dense connections, plays a key role in providing special advantages. Main advantages of such deep networks over ResNet [9] and InceptionNet [10] are in terms of better understanding of the model, complexity reduction, less requirement of parameters, lower computational and power consumption, and higher accuracy.

Generally, a DenseNet contains blocks which are combinations of multiple convolution layers of dense connectivity. Our DenseNet model contains only one dense block with 16 convolutional layers in it and other components of the model are same as [8] except the transition block which we do not use. Similarly, batch normalization, relu activation, global average pooling, fully connected neural networks and Softmax layers are the essential layers in the proposed model.

*3.4.1. Training and Inference*
To train an efficient deep learning model, the tuning of hyperparameters plays key role. The proposed model has the common hyperparameters including the growth-rate (K), dropout, depth of network, number of dense blocks, learning rate, weight decay, optimizer, momentum, epochs, batch size, and weight regularization. To start training process on 30951 image patches of entities with two classes, we use K=12, dropout=0.2, depth of network =7, epochs=40, number of dense blocks=1, learning rate=$1\times10^{-3}$, weight decay=$1\times10^{-4}$ and batch size=64. To optimize the weights of the model, we use Adam optimizer which is a combination of stochastic gradient descent optimizer having a momentum of 0.9 and RMSprop. In our implementation of DenseNet we set categorical cross entropy as a loss function.

We selected major hypermeters different from by default setting of DenseNet [8] and similar to a model in [11]. The main reason behind this is that we trained our model on a usual core i3 laptop instead of highly computational capable GPUs. Our training machine has 4 CPUs with 2.5 GHz speed and a 4GB RAM. Similarly, during training process, our proposed model converges with top results in term of a loss=0.13 and accuracy=0.96. The inference was conducted on 10,318 test patches which are not used in training.

## 4. RESULTS
To test the accuracy of our anomaly detection system, the trained cascaded models are evaluated on test images of examination hall dataset.

Figure 2 shows the confusion matrix of our experiments on 10318 test image patches. Here, normal class means that participants are not involved in any cheating or illegal activities and abnormal class is the opposite of it. This figure indicates that our method can detect abnormal entities with an accuracy of 78%.

Results of Table 1 shows that the normal class is almost 8 times greater than abnormal class therefore, the problem of class imbalance can be the reason behind less sensitivity of the abnormal class. Moreover, the overall results of our method are quite satisfactory with a 96% precision, sensitivity, and F1 score. The proposed method achieves a 95.88% overall accuracy for both classes.

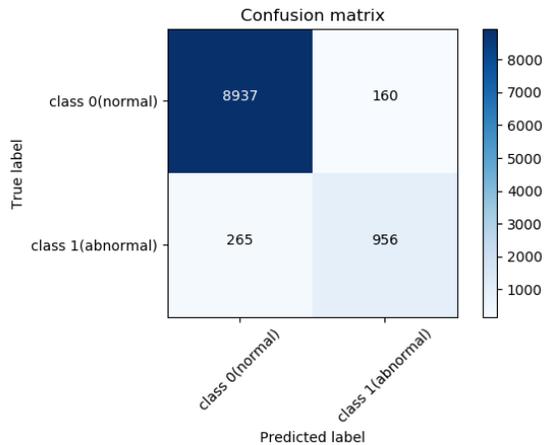

**Figure 2** Confusion matrix of our proposed method.

**Table 1** Quantitative results of our method.

|  | Precision | Sensitivity | $F_1$-Score | Images |
|---|---|---|---|---|
| 0(normal) | 0.97 | 0.98 | 0.98 | 9097 |
| 1(abnormal) | 0.86 | 0.78 | 0.82 | 1221 |
| avg / total | 0.96 | 0.96 | 0.96 | 10318 |

Figure 3 represents the predictions of our model on our dataset. It can be observed that our method can detect even the very subtle anomalous behavioral cues such as slight head angle, gaze directions or persons trying to talk.

## 5. DISCUSSION

Table 1 and Fig. 2 show that our method have an imperfect precision score for abnormal participants. The major factors which effect accuracy and precision may be related to the depth of the models. The precision of our method can be increased by using GPUs that will help to increase number of dense block and depth of network. Moreover, precision score can also be improved by solving the problem of class imbalance. For instance, augmentation or new data samples for abnormal class can be a possible technique to balance the dataset. The overall results show that our cascade of the two deep networks i.e. the 2D pose estimation and DenseNet gives our system the ability to understand and handle complex features of entities like movements of heads, direction of eyes and face.

## 6. CONCULUSION

This paper proposed a novel computer vision method to detect anomalous entities in an indoor examination hall environment. Our method explored a cascade of deep convolutional neural networks models including OpenPose and DenseNet for the task. Our system achieved an overall accuracy of 95.88% after training a lightweight but robust DenseNet on a Quadcore CPU with 4GB RAM.

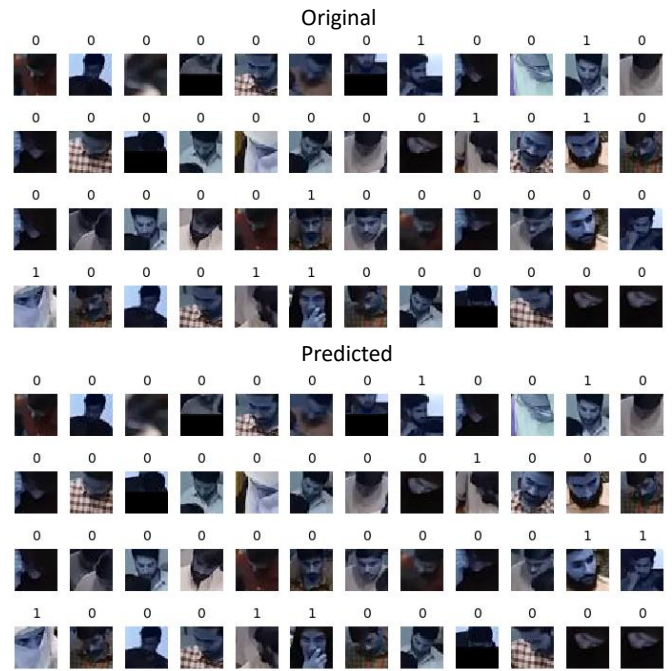

**Figure 3** Examples of predictions made by our method.